\pdfoutput=1
\documentclass[11pt]{article}

\usepackage{acl}
\usepackage[utf8]{inputenc}
\usepackage[T2A,T1]{fontenc}
\usepackage[russian,english]{babel}
\usepackage{graphicx}

\usepackage{times}
\usepackage{booktabs}
\usepackage{latexsym}
\usepackage{hyperref}
\usepackage{lingmacros}

\usepackage{microtype}

\newcommand{\rusword}[1] {`\foreignlanguage{russian}{#1}'}

\title{RuDSI: graph-based word sense induction dataset for Russian}

 \author{Anna Aksenova \\
 National Research University \\ Higher School of Economics \\ Russia \\
 {\tt a.aksenova@hse.ru} \\\And
 Ekaterina Gavrishina, Elisey Rykov \\ National Research University \\ Higher School of Economics \\ Russia \\
 {\tt eigavrishina@edu.hse.ru} \\ {\tt esrykov@edu.hse.ru} 
 \\\And
 Andrey Kutuzov \\
 University of Oslo\\ Norway \\
 {\tt andreku@ifi.uio.no} \\}

\begin{document}

\maketitle

\begin{abstract}

We present RuDSI, a new benchmark for word sense induction (WSI) in Russian. The dataset was created using manual annotation and semi-automatic clustering of Word Usage Graphs (WUGs). Unlike prior WSI datasets for Russian, RuDSI is completely data-driven (based on texts from Russian National Corpus), with no external word senses imposed on annotators. Depending on the parameters of graph clustering, different derivative datasets can be produced from raw annotation. We report the performance that several baseline WSI methods obtain on RuDSI and discuss possibilities for improving these scores.

\end{abstract}

\section{Introduction}

Word sense induction (WSI) is among the most challenging problems in computational linguistics. The difficulty lies not only in the character of the task itself but also in the lack of datasets properly designed for it. We have developed such a dataset for the Russian language by means of manual annotation and clustering of the obtained senses. We dub it \textit{Russian Data-driven Sense Induction} dataset (RuDSI)\footnote{\url{https://github.com/kategavrishina/RuDSI}}. Its annotation was based on so-called Word Usage Graphs (WUGs), where word usages in context are nodes connected by edges with weights corresponding to semantic proximity \citep{schlechtweg-etal-2020-semeval}. This workflow has been already used to create diachronic semantic change datasets for Russian \citep{rodina2020rusemshift, kutuzov2021three}, but it is the first time it is employed for designing synchronic WSI benchmarks.

Graphs representing semantic relations between word usages were crucial for the creation of RuDSI. Communities or clusters induced from these graphs correspond to lexical senses; the number and composition of clusters for each word depends not only on human annotation, but also on the particular clustering procedure. Since we provide raw annotators' judgments, other researchers can apply their preferred graph processing techniques and obtain slightly different sense assignments.

The rest of the paper is organized as follows. In Section~\ref{sec:related}, we talk about the WSI datasets created earlier and their limitations. In Section~\ref{sec:dataset}, we present and analyze RuDSI and describe our annotation workflow. In Section ~\ref{sec:robustness}, we show how graph clustering parameters affect the dataset. Section~\ref{sec:baselines} reports the performance of several baseline WSI methods. In Section~\ref{sec:audience}, we describe to whom and how RuDSI will be useful.\footnote{This research was supported in part through computational resources of HPC facilities at HSE University \cite{Kostenetskiy_2021}.}

\section{Related work}\label{sec:related}

In this section, we give a brief overview of word sense induction datasets for English developed as a part of SemEval competition, take a look at RUSSE'18 dataset and discuss the approaches towards WSI dataset creation. 

\subsection{SemEval datasets}

Existing sense-annotated corpora like SemCor \cite{miller1993semantic} allow for building competitive word sense disambiguation (WSD) models since they provide sufficient amount of training data. However, the major problem of such sources is the fact that word sense inventories vary depending on text domain and time period. Thus, WSD models are never universal. To solve this issue, word sense induction task was created. WSI systems aim to infer word senses from the given corpus.

In 2010, a WSI dataset was introduced during the SemEval competition \cite{manandhar-etal-2010-semeval}. Compared to SemEval 2007 \cite{agirre-soroa-2007-semeval}, it was more balanced in terms of nouns and verbs distribution (50 verbs and 50 nouns in English). The main difference was in the evaluation procedure. The authors assumed that although WSI task is unsupervised, evaluating the methods on unseen test set of contexts would be more realistic. Different metrics for the clustering quality evaluation were inspected (V-measure, paired F-score) and all of them turned to be biased by number of senses predicted by WSI algorithms.

In 2013, another task setting was suggested by \citet{jurgens-klapaftis-2013-semeval}. They claimed that there are contexts where multiple sense tags might be used. Therefore, the setup required predicting the weighted distribution of word senses for each context, i.e., perform \textit{graded} word sense induction. To evaluate this task, two novel measures were introduced: fuzzy B-Cubed and fuzzy normalized mutual information. We should emphasize that our RuDSI dataset is aimed to test systems for \textit{non-graded} word sense induction, although it could be transformed into graded setup (see Section \ref{section:graph_clus}). 

\subsection{Russian WSI datasets}

Despite the fact that word sense induction task was well-developed for English, there were no manually annotated data for Russian until recently. In the last years, the interest to WSI and WSD tasks in Russian has increased due to the appearance of the first Russian WSI dataset. It was created as a part of RUSSE-18 shared task \cite{Panchenko:18:dialogue} and contains three subsets: 
\begin{enumerate}
    \item \textbf{wiki-wiki} (automatically extracted examples and senses from Wikipedia articles, mainly homonyms and homographs)
    \item \textbf{bts-rnc} (examples from the Russian National Corpus (RNC), labeled with senses from the `Big Explanatory dictionary')
    \item \textbf{active-dict} (examples and senses from the `Active dictionary of the Russian language' by Yuri Apresjan \cite{apresjan2014active})
\end{enumerate}

The training sections contained a total of about 17 thousand contexts. The key metric for the competition was Adjusted Rand Index (ARI) score \cite{hubert1985comparing}. The Rand Index calculates the similarity between two clusterings by counting object pairs that were assigned the same or different clusters in golden labeling and in predictions. ARI adds adjustment for chance and gives score close to 0 for random labeling and 1 for identical clusterings. When the clustering is worse than random, ARI is negative.

\subsection{Limitations of previous datasets}

Unfortunately, RUSSE-18 shared task data has a number of significant limitations. Linguistically, it includes homonyms, polysemous words and homographs, which does not correspond to the original WSI task setting: inducing senses of lexemes with the same set of word forms. In addition, some of the contexts in RUSSE-18 are noisy: there are cases where the target word is actually a root of a composite or a derivation (e.g., \rusword{луковица} \textit{bulb} is suggested as one of the words in context set for target word \rusword{лук} \textit{onion/bow}). The key issue is that word sense cannot be induced in these cases since derivations are mostly non-compositional and do not necessarily maintain the ambiguity relations of parent word. Finally, none of the target words of RUSSE-18 are monosemous, hence the dataset does not test WSI systems for polysemy detection, which is a critical issue in terms of developing a universal algorithm. 

All the datasets for both Russian and English SemEval discussed above were automatically or manually tagged with dictionary-based sense inventories. We believe that it might be more realistic to derive word sense inventories for WSI pipelines evaluation not from linguistic sources, but directly from corpora, since the sets of senses vary in different corpora and domains \cite{kilgarriff1997don}.

\subsection{Graph-based WSI datasets}

A possible solution comes from combining word-in-context disambiguation and graph clustering. Conceptualization of semantic relationships as graphs empowered the approaches that represent the ambiguous lexeme as a central node in graph where nodes are the words and edge weights represent the measure of association between those words. \citet{hope2013maxmax} suggests calculating edge weights as a frequency measure for word co-occurrence similarity: the more similar are the contexts of the node lexemes, the higher will be the edge weight bridging them. Such co-occurence graphs are calculated automatically. The similarity networks are afterwards clustered to induce word senses \citep{hope2013maxmax, sherstuk}.

\citet{mccarthy2016word} highlighted the problem of using fixed sets of senses for word sense inventory representation. Graphs used in the paper were derived from word in context disambiguation annotation. They suggested treating annotators' judgements as graph edges and investigated different clusterability measures of such graphs. 

Graph clustering has been successfully employed in annotating datasets for semantic change detection task \cite{schlechtweg-etal-2020-semeval, schlechtweg-etal-2021-dwug}. The annotation process is essentially word-in-context disambiguation: the annotators have to decide whether a pair of sentences represent the same target word sense or not. The annotation forms a \textit{word usage graph} combining the uses from each pair of word contexts, where the nodes are the contexts themselves (sentences), and edges are weighted with the medians of annotators' judgments for a particular pair. Then, using correlation clustering, the graph is separated into clusters (communities of nodes) that correspond to the senses. The method is simple yet quite efficient as the annotators do not assign sense labels directly and the resulting clusters represent a set of data-driven senses\footnote{As opposed to dictionary-based senses, since the obtained senses are not taken from any resources, they are the result of automatic clustering.}. Such a method does not only represent the relations between word usages, but also allows for choosing the granularity of the final word sense inventory. Moreover, the resulting senses are derived from data and not biased by lexicographic information; also, the number of clusters is determined automatically \cite{schlechtweg-etal-2021-dwug}.

\section{RuDSI dataset} \label{sec:dataset}

\subsection{Target words selection}
To create RuDSI, it was first necessary to select a limited number of target words for further manual annotation. As we aimed at having words of different degree of polysemy presented in the final dataset, we extracted the total number of senses for each word in three distinct resources: Russian National Corpus (RNC)\footnote{\url{https://ruscorpora.ru}; in particular, we used the RNC semantic markup \citep{rahilina2009zadachi} which includes parts of speech and semantic classes for a large number of lexemes (for example, \textit{fruit/food} for the word `apricot').}, representative collection of texts in Russian with linguistic annotation; Wiktionary\footnote{\url{http://www.wiktionary.org}}, web-based free dictionary; and RuWordNet\citep{loukachevitch2016creating}, a thesaurus of the Russian language created in the format of English WordNet \citep{miller1995wordnet}. All non-noun words were discarded from this set. 

Since the purpose of the annotation was to create a dataset with a balanced number of mono- and polysemous lexemes, we selected eight most frequent words (according to the dictionary by \citet{lyashevskaya2009}) in each of three groups: words with one sense, words with 2-4 senses (moderately polysemous), words with five or more senses (highly polysemous).
The value of eight was chosen because of our limitations on the volume of annotation. The final number of senses was calculated as the average\footnote{The average was preferable to minimum and maximum, since they would give more weight to one of the resources: in Wiktionary, words usually have few senses (1-2), but in RNC, same words can have a lot more senses (6 on average).} between RNC, Wiktionary and RuWordNet for each target word.
Note that we did not consider these values as any sort of a gold standard, and they did not affect our human judgements in any way: annotators were not aware about the polysemy groups which the target words belonged to.

Thus, 24 target nouns were prepared for the annotation. For each word from the resulting set, 35 sentences containing this word were randomly sampled from the RNC. Next, annotators were given pairs of these sentences to estimate the relatedness of target word senses between each element in the pair.

\subsection{Annotation}

\begin{figure*}[ht]
\centering
\includegraphics[width=\textwidth]{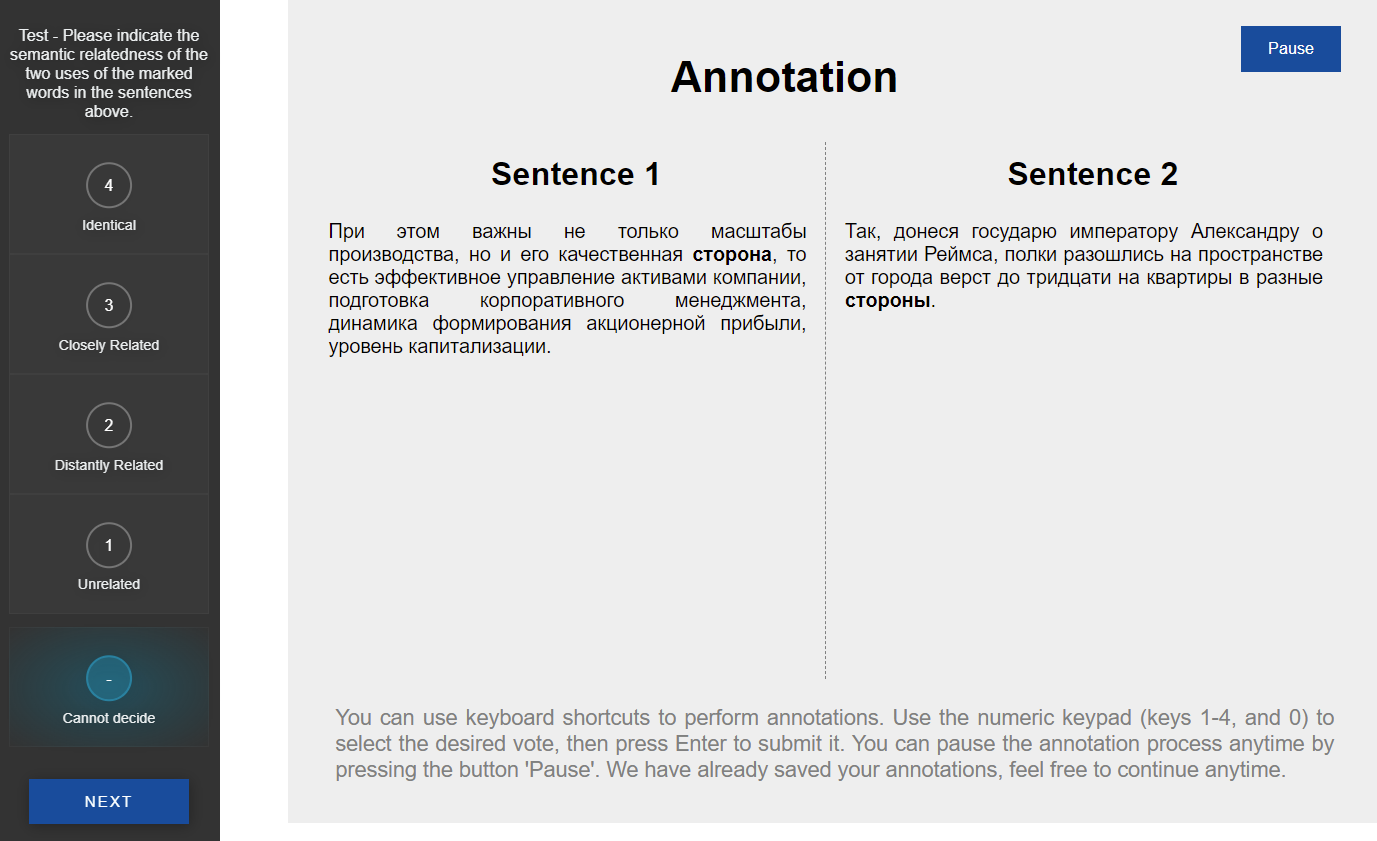}
\caption{Example of the word \rusword{сторона} (\textit{side, direction}) annotation in the DURel interface.}
\label{figure:durel}
\end{figure*}

The annotation was performed using the DURel web service\footnote{\url{https://durel.ims.uni-stuttgart.de}}, which allows to annotate pairs of contexts for each word from the loaded sample. At each step of the annotation, a human is offered a pair of sentences to judge. For each pair, the columns `Sentence 1' and `Sentence 2' are presented with contexts containing the target word, which is highlighted in bold. The task is to assess how close in meaning the occurrences of the target word are in the two presented sentences on the scale from 1 (Unrelated) to 4 (Identical). The scores of 2 (Distantly Related) and 3 (Closely Related) are more subjective. In general, the 2 rating is for the uses that have different senses, but are somewhat related, and the 3 rating is for the cases when two uses have the same sense with some variation. So, a score of 1 is implied in the following example with the target word \rusword{сторона} which is presented in the Figure \ref{figure:durel}, indicating that there is no connection between the senses (direct and figurative meaning of the lexeme):

\eenumsentence{
\item \rusword{При этом важны не только масштабы производства, но и его качественная сторона, то есть эффективное управление активами...} (the meaning of \textit{`component, element'})\\
\textit{At the same time, not only the scale of production is important, but also its qualitative \textbf{side}, that is, effective asset management...}

\item \rusword{Так, донеся государю императору Александру о занятии Реймса, полки разошлись на пространстве от города вёрст до тридцати на квартиры в разные стороны.} (the meaning of \textit{`space, direction'})\\
\textit{So, having informed the Emperor Alexander about the occupation of Reims, the regiments dispersed in the space from the city to thirty versts to apartments in different \textbf{directions}.}}

The next example presents the case of two uses with identical senses for the word \rusword{день} (\textit{day}) requiring the score of 4:

\eenumsentence{
\item \rusword{Вещи не были еще расставлены, рамы были частью без стекол, частью с остатками расколотых, и (был дождливый день) с потолка текло.}\\
\textit{Things were not yet arranged, the window frames were partly without glass, partly with the remains of splintered ones, and (it was a rainy \textbf{day}) the ceiling was flowing.}

\item \rusword{Каждый день с раннего утра до обеда и с обеда до вечера я занят был работою или в доме, или в саду, или в огороде}\\
\textit{Every \textbf{day}, from early morning to lunch and from lunch to evening, I was busy working either in the house, or in the garden, or in the vegetable garden.}}

For each of the 24 words, as mentioned earlier, 35 sentences were sampled from the RNC. The DURel platform automatically generated random sentence pairs, and at the first stage of our workflow, 180 pairs were annotated for each target word\footnote{Annotation was performed by a subset of the authors of the article as native Russian speakers.}. As a result, 24 separate word usage graphs with 35 nodes each were obtained.

\subsection{Aggregation of senses via graph clustering}
\label{section:graph_clus}

Clustering of the sentences obtained as a result of the annotation for each lexeme was performed using the pipeline from \citep{schlechtweg-etal-2021-dwug} based on the variation of correlation clustering \cite{bansal2004correlation, schlechtweg-etal-2020-semeval}.
The DURel relatedness scale from 1 to 4 was derived from continuum of semantic proximity \citep{blank1997}: Homonymy - Proximity - Context Variance - Identity. Based on the continuum, the authors rescaled the annotators' judgements for clusterization to represent the idea of usage pairs with 1 and 2 scores belonging to different senses, and with 3 and 4 scores — to the same sense.
For this purpose they created the \textit{threshold} parameter which was used to calculate the resulting edge weight: \(W'(e) = W(e) - threshold\), and equated it to 2.5 (e.g., a score of 1 became -1.5).
The division into clusters is based on the similarity between the target word senses within the sentences in a pair: clustering algorithm minimizes the sum of positive edge weights (3 and 4 scores in the original) across clusters and the sum of negative edge weights (1 and 2 scores) within clusters.
Correlation clustering yields only one cluster label for a node (sentence), but by replacing it with a fuzzy graph clustering algorithm like the one in \cite{PENG202138}, it is possible to come up with a graded variation of RuDSI. 

As a result of clustering, sense clusters were obtained, which contain examples for each target word, labeled with sense number and connected by edges (the edge weight depends on the number and values of annotators' judgements).

After the first round of annotation, we analyzed the number of \textit{uncompared clusters} — those clusters whose sentences have never been compared in the process of annotation. The existence of uncompared clusters indicates that the graph is not connected enough. We decided that for the five words with the number of uncompared clusters exceeding the average (2.75) additional annotation is required. After the second annotation round (60 extra pairs of sentences for each of five words) there were still four words left for which the number of uncompared clusters has remained almost the same and still exceeded the average number. For these words, sentences from the corresponding clusters were manually selected, organized into pairs and annotated following our regular workflow. After all the annotation rounds, the number of uncompared clusters is not more than two for any target word, and the average number of annotated sentence pairs per word is 215.

Initially, we got a large number of singleton clusters (1.13 on average across words). These are clusters containing only one node (context, usage example). They may appear when the target word is used in a specific context, for example, in an idiomatic expression. Singleton clusters are problematic, since in these cases it is difficult to tell legitimate exotic senses from clustering errors. We planned to filter them out in one of the following ways: not to consider examples from singleton clusters or to attach singleton examples to the largest cluster of a particular word, reducing the total number of senses. However, after reviewing the clusters manually, we noticed that in some lexemes singleton clusters can be aggregated with a larger one, but not with the largest one, and in other lexemes singleton clusters, on the contrary, express a very specific idiomatic expression that can neither be attached to another cluster nor removed from the sample without loss of representative power. So, we decided to leave the singleton clusters untouched and did not filter them out.

\begin{figure}[ht]
\centering
\includegraphics[width=\columnwidth]{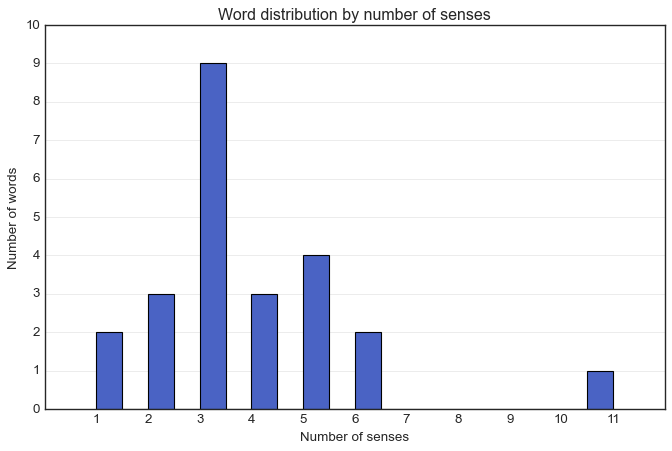}
\caption{Word distribution by the number of senses obtained in RuDSI.}
\label{figure:distribution}
\end{figure}

Figure~\ref{figure:distribution} shows the distribution of the number of senses for the target words yielded by the annotation procedure (per-word numbers can be found in the Appendix). As can be seen, most words tend to end up having 3-5 senses.

\subsection{Important statistics}

Based on the results of clustering, we computed some statistics presented in this subsection. In particular, the ratios of words by the number of senses was calculated. As it turned out, RuDSI contains 8.3\% of monosemous words, 62.5\% of words with 2-4 senses (moderately polysemous), and 29.2\% of words with five or more senses (highly polysemous). Note that these values are different from the original percentages obtained from our linguistics sources. This is expected, since our senses are fully data-driven.

It was also interesting to consider the correlation of these `data-driven' sense numbers and the degree of lexical polysemy yielded by the RNC, Wiktionary and RuWordNet, on which we relied during the selection of the target words. The Table \ref{table:correlation} shows Spearman correlation between the number of clusters in the RuDSI word usage graphs and the number of senses in the sources mentioned above. `Mean number of senses' is the average between the RNC, Wiktionary and RuWordNet. All the correlations are strong and significant at $p=0.05$: that is, the resulting clusters based on data-driven sense induction roughly correspond to sense numbers from external linguistic sources.

\begin{table}[ht]
\begin{center}
\resizebox{.4\textwidth}{!}{
\begin{tabular}{l l l}
\toprule
\textbf{Source} & \textbf{\textit{Spearman $\rho$}} & \textbf{\textit{p value}}\\
\toprule
RNC   & 0.84 & 0.000 \\ 
Wiktionary  & 0.43 & 0.034 \\
RuWordNet    & 0.73 & 0.000 \\
Mean number of senses & 0.90 & 0.000 \\
\bottomrule
\end{tabular}
}
\caption{Correlation of the word sense numbers between RuDSI and other resources.}
\label{table:correlation}
\end{center}
\end{table}

In addition, we calculated the Spearman correlation between the number of senses in RuDSI and the target word frequencies from the \citet{lyashevskaya2009} dictionary (based on the RNC). Its value is 0.53 ($p=0.007$). Therefore, the number of word senses in RuDSI is significantly correlated with word frequencies in the RNC. This is expected, since it is known that frequent words tend to be more polysemous \cite{zipf1945meaning, hernandez2016testing}. It also means than in many cases it is possible to predict the number of RuDSI senses for a word by looking at its RNC frequency.

\subsection{Format and technical details}

As a result of the steps described above, each example sentence (usage) for each target word was assigned an index of the cluster to which it belongs. We aggregated this data in order to compile a dataset in a format similar to RUSSE-18 \cite{Panchenko:18:dialogue}. The structure of the RuDSI dataset is presented in the Table \ref{table:rudsi}: word, context (sentence), positions of the word in the context and the gold identifier of the cluster (sense).

\begin{table}[ht]
\begin{center}
\resizebox{0.5\textwidth}{!}{
\begin{tabular}{l l l l}
\toprule
\textbf{word} & \textbf{context} & \textbf{positions} & \textbf{sense\_id} \\
\toprule
\rusword{тысяча} & \rusword{...тысяча пятьсот запорожцев...} & 76-82 &	0 \\
\rusword{тысяча} & \rusword{...вмещает 12 тысяч зрителей.} & 34-39 &	0 \\
\rusword{тысяча} & \rusword{...около 5 тыс. вагонов...} &	49-52 & 0 \\
\rusword{тысяча} & \rusword{...ну, сотни тысяч.} &	34-39 & 0 \\
\rusword{тысяча} & \rusword{...на пытки тысячи ни в чем...} & 28-34 & 1 \\
\bottomrule
\end{tabular}
}
\caption{RuDSI dataset sample for the word \rusword{тысяча} (\textit{thousand}).}
\label{table:rudsi}
\end{center}
\end{table}

We encourage evaluating state-of-the-art WSI approaches with RuDSI, this is why it was important for the texts in the dataset to not exceed 512 tokens in length. The maximum sequence length is always added to the Transformers architecture models due to the attention layers, which are quadratically scaled with increasing sequence length. 512 tokens is the popular maximum sequence length, which was first specified in BERT. The only sentence in RuDSI (out of 840) which has been longer than this value has been truncated to 512 tokens.

\section{Robustness of clustering}\label{sec:robustness}

In order to verify the stability of clustering algorithm we experimented with changing the default hyperparameters and analyzed the resulting data in comparison with the default sense clusters presented in RuDSI.
In the pipeline \citep{schlechtweg-etal-2021-dwug}, there were two parameters that could affect the obtained clusters: the threshold used to rescale the annotators' judgements and the number of clustering iterations. The threshold parameter was previously described in \ref{section:graph_clus}: it affects the resulting weights on the graph edges. Originally, the threshold was 2.5 causing 1 and 2 scores (`Unrelated' and `Distantly Related') to transform to negative values, and 3 and 4 scores (`Closely Related' and `Identical') to remain positive to represent the contrast between different senses and the same sense of the word. We reviewed two other options: the threshold equaled to 1.5 (so that a score of 1 became negative (-0.5) and contrasted with 2, 3 and 4 scores that were matched to 0.5, 1.5 and 2.5 respectively) and equaled to 3.5 (1, 2 and 3 scores were opposed to a score of 4; only the sentences marked us `Identical' were considered as containing the same sense of the word).

The number of clustering iterations (`iters' parameter) stands for the number of passes through the same graph given that the input graph is the result of the previous iteration. Each pass performs the clustering algorithm and minimizes the loss of the obtained clusters.

In Table \ref{table:clustering}, are presented the mean and standard deviation of ARI score among words between the default clustering and clusterings with modified hyperparameters. We can conclude that the number of iterations does not greatly affect the resulting clusters, even as a result of a single iteration (`iters' = 1) approximately the same clustering is obtained. However the threshold parameter strongly influences the obtained clusters as it reforms the original idea of similarity of different judgements during the annotation.

\begin{table}[ht]
\begin{center}
\resizebox{.4\textwidth}{!}{
\begin{tabular}{llll}
\toprule
\textbf{iters} & \textbf{threshold} &  \textbf{\textit{Mean ARI}} &    \textbf{\textit{SD ARI}} \\
\toprule
  5 &       2.5 &        -- &        --  \\
\midrule
  5 &       1.5 &  0.12 &  0.29 \\
  5 &       3.5 &  0.27 &  0.26 \\
\midrule
  1 &       2.5 &  0.95 &  0.13 \\
  3 &       2.5 &  0.95 &  0.10 \\
  4 &       2.5 &    0.95 &   0.11 \\
  6 &       2.5 &  0.94 &   0.13 \\
\bottomrule
\end{tabular}
}
\caption{Similarity (by ARI) of the default RuDSI clustering and clusterings obtained by changing hyperparameters. `SD' stands for standard deviation.}
\label{table:clustering}
\end{center}
\end{table}

We also examined the change in the number of singleton clusters depending on clustering hyperparameters. Similarly, the threshold parameter has a much stronger effect than the number of iterations. The threshold of 1.5 causes merging of most senses into one cluster (sense) and separation of the minimal number of singleton clusters (0.13 on average). In turn, the threshold of 3.5 generates division into a larger number of clusters most of which are singleton clusters (6.33 on average). Notably, the iterations parameter is inversely proportional to the number of singleton clusters: the more iterations, the more singleton clusters are attached to larger clusters (the more senses are considered the same). A summary of singleton clusters analysis can be found in Table \ref{table:singleton}.

\begin{table}[ht]
\begin{center}
\resizebox{.4\textwidth}{!}{
\begin{tabular}{llll}
\toprule
\textbf{iters} & \textbf{threshold} & \textbf{\textit{\# Singletons}} &  \textbf{\textit{SD}} \\
\toprule
  5 &       2.5 &  1.13 &  0.74 \\
\midrule
  5 &       1.5 &    0.13 &  0.45 \\
  5 &       3.5 &   6.33 &  4.43 \\
\midrule
  1 &       2.5 &  1.21 &  0.83 \\
  3 &       2.5 & 1.13 &  0.8 \\
  4 &       2.5 &     1.13 &  0.95 \\
  6 &       2.5 &  1.08 &  0.88 \\
\bottomrule
\end{tabular}
}
\caption{Statistics for singleton clusters in the default RuDSI clustering and clusterings obtained by changing hyperparameters. `Singletons' is the average number of singleton clusters among words. `SD' stands for standard deviation.}
\label{table:singleton}
\end{center}
\end{table}

\section{Baseline WSI methods performance} \label{sec:baselines}
In this section, we show how the existing WSI methods perform on RuDSI. We deliberately did not experiment with the state-of-the-art lexical substitution method \cite{amrami2019towards}. The goal is to report the results of the baseline approaches, leaving more advanced methods for future research.

\subsection{Naive baselines}
Two naive baselines were implemented for WSI problem solution, namely assignment of the same sense for all target words, and a random choice of two senses for each target word.

\subsection{Birch}

Next, we applied more advanced embedding-based approaches. One of the methods top-rated in the RUSSE-18 shared task is static embeddings clustering \citep{Panchenko:18:dialogue}. After testing different clustering algorithms, we settled on Birch \citep{10.1145/233269.233324}, which provided the best results. We used the following pipeline: first, we calculated sentence embeddings as an average over word embeddings for each context, second, all embeddings within each target context were divided into two clusters. For word embedding extraction we used \textit{ruwikiruscorpora-func\_upos\_skipgram\_300\_5\_2019} Word2Vec model trained on RNC and Wikipedia from the RusVectores web service \citep{KutuzovKuzmenko2017}.

\subsection{Jamsic}

Jamsic method was also included in the list of the best systems in the RUSSE-18 shared task description paper \citep{Panchenko:18:dialogue}. Using the Word2Vec model specified above, the nearest neighbor for each target word is extracted. The embedding of this word represents the first sense of the target word. Then this embedding is subtracted from the embedding of the target word and the embedding of the second sense is obtained. Finally, we get an average embedding for each sentence and determine to which sense it is closer by cosine similarity. This method works with one word sense and its nearest one, so it always distributes contexts into only two senses.

\subsection{Egvi}

This is a relatively new approach that has successfully proved itself in solving the WSI problem for different languages. For this method, we used Russian sense inventories pre-generated by processing ego graphs \citep{logacheva-etal-2020-word}, and for each target word we received an average embedding of each sense from sense inventories. For word embeddings, we used the same \textit{ruwikiruscorpora-func\_upos\_skipgram\_300\_5\_2019} model. Then we removed the target word from RuDSI contexts, received average word embeddings and clustered them with the KMeans algorithm, passing embeddings of values from sense inventories as cluster centers. The parameter of number of clusters for each target word was equal to number of senses in sense inventories for this word.

\subsection{BERT KMeans}

BERT-based embeddings proved to be efficient in solving RUSSE-18 too \citep{slapoguzov2021word}. We took the \textit{sbert\_large\_nlu\_ru} model\footnote{\url{https://huggingface.co/sberbank-ai/sbert_large_nlu_ru}} as a feature extractor and used token embeddings from its last layer. For calculating the representation of words split during tokenization, mean pooling was used. Word vectors were clustered by the KMeans algorithm into two senses. 

We also tried to do KMeans clustering of BERT embeddings by taking the number of clusters from Egvi sense inventories.

\subsection{Results}
The mean and standard deviation of ARI score among words are presented in Table \ref{performance_wsi}. The ARI metric takes into account randomness when clustering, so the ARI of the Random sense method is 0.0. The approaches that became the best in the RUSSE-18 shared task do not gain values higher than 0.05 on RuDSI. The simplest One sense baseline is better than BERT clustering. Arguably, the low BERT results are caused by the number of clusters parameter of the KMeans algorithm, which was equal to 2, while only two target words (out of 24) actually had two senses. Egvi algorithm proved to be the best. This method was based on the pre-generated sense inventories, in which the number of senses often was identical to RuDSI, so it worked better than BERT KMeans. The number of Egvi senses improved the quality of clustering of BERT embeddings, but not enough to exceed the native Egvi. 

For comparison, the table shows the results of the methods on the RUSSE-18 dataset. Due to a number of limitations described earlier, a wiki-wiki dataset was taken for comparison. It is noticeable that the values of the ARI metric for the basic methods on wiki-wiki are much higher.

We also found no correlation between the density of the word graph and the values of the ARI metric, with the exception of the Jamsic method, for which the correlation results are significant at a significance level of $p=0.05$.

\begin{table}[h!]
\begin{center}
\resizebox{0.5\textwidth}{!}{
\begin{tabular}{@{}lllll@{}}
\toprule
\multicolumn{1}{c}{{\textbf{Method}}} & \multicolumn{2}{c}{\textbf{RuDSI}} & \multicolumn{2}{c}{\textbf{RUSSE}} \\ \cmidrule(l){2-5} 
\multicolumn{1}{c}{}                                 & Mean ARI            & SD ARI       & Mean ARI          & SD ARI         \\ \midrule
One sense                                            & 0.08                & 0.28         & 0.00              & 0.00           \\
Random sense                                         & 0.00                & 0.00         & 0.01              & 0.00           \\
Birch                                                & 0.03                & 0.14         & 0.93              & 0.10           \\
Jamsic                                               & 0.04                & 0.10         & 0.58              & 0.47           \\
BERT KMeans                                          & 0.03                & 0.14         & 0.85              & 0.06           \\
BERT KMeans + Egvi                                   & 0.08                & 0.31         & 0.64              & 0.31           \\
Egvi                                                 & \textbf{0.17}       & 0.22         & 0.59              & 0.16           \\ \bottomrule
\end{tabular}
}
\caption{Performance of WSI methods on RuDSI and RUSSE. `SD' stands for standard deviation.}
\label{performance_wsi}
\end{center}
\end{table}

\section{Intended RuDSI audience}\label{sec:audience}
Our vision is that RuDSI might be of use for three different communities.

\begin{enumerate}
    \item Researchers analyzing NLP systems in terms their WSD and  WSI abilities for Russian. It is especially important for evaluating contextualized language models trained on large-scale corpora using deep neural architectures, from RNNs to  Transformers and beyond. RussianSuperGLUE benchmark \cite{shavrina-etal-2020-russiansuperglue} already includes the RUSSE dataset (cast as a binary classification task). We believe RuDSI might be a useful addition, representing a more difficult task related to lexical senses. As was shown in \ref{sec:baselines}, it cannot be solved with trivial baselines \cite{iazykova2021unreasonable}, which makes it an interesting NLP challenge.
    \item Graph theory researchers and all those interested in applications of graphs to real world tasks. Word usage graphs we are providing are representative of contextual semantic similarity judgments by humans. These graphs can be processed and clustered in different ways, yielding different `views' of Russian word sense inventories. In addition, the properties of word usage graphs themselves can bring new insights for both graph theory and Russian linguistics.
    \item Finally, our work on RuDSI is a part of a larger project of implementing WSI features into the RNC web interface. RuDSI is based on RNC data, so it will be used to evaluate various WSI solutions and choose the best one. Thus, it is going to be directly or indirectly used by the large RNC audience, consisting of both linguists and general population. 
\end{enumerate}

\section{Conclusion}
We have presented RuDSI, a novel graph-based word sense induction dataset for Russian, obtained by clustering word usage graphs produced by human annotation. It includes words with different degrees of polysemy (monosemous, moderately polysemous and highly polysemous words). The sense inventories are generated in a completely data-driven way as well. Importantly, depending on what graph processing workflow is used, slightly different datasets can be produced from the same raw RuDSI human judgments.

We report the RuDSI performance for only the simplest and most basic approaches to WSI, so a possible future work would be to apply some more advanced methods to it. Also we have considered only nouns, so it would be interesting to experiment with other parts of speech as well (this will require a new round of annotation). Since most of the target words in RuDSI have 3-5 senses, the addition of highly polysemous words may become another future improvement. In addition, it would be beneficial to extend the list of contexts for each word, however extra annotation would be required.

\bibliography{bibliography}
\bibliographystyle{acl_natbib}

\appendix

\section{Annotation interface}

\section{Word usage graph}

\begin{figure*}[ht]
\centering
\includegraphics[width=\textwidth]{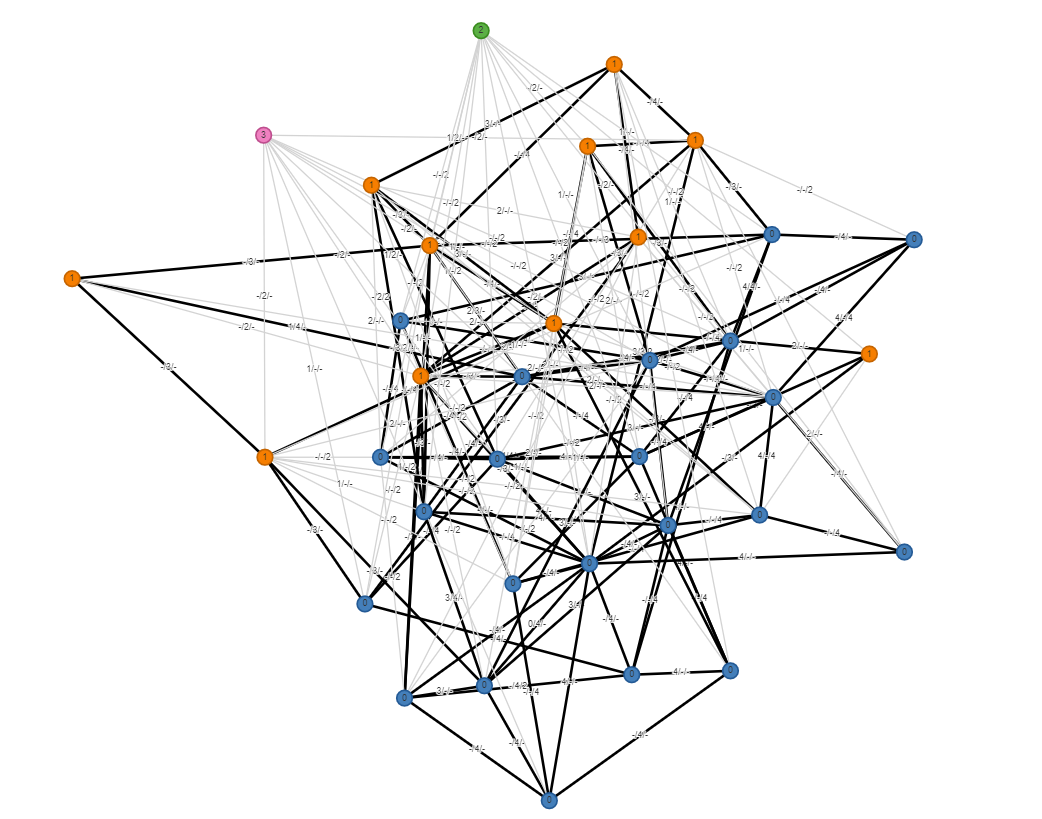}
\caption{Word usage graph for the word \rusword{голова} (\textit{head}) as a result of clustering (four clusters, marked with node color).}
\label{figure:graph}
\end{figure*}

\section{Detailed performance}

\begin{table*}[ht]
\resizebox{\textwidth}{!}{
\begin{tabular}{lllllllll}
\toprule
Word &
  \multicolumn{1}{l}{One sense} &
  \multicolumn{1}{l}{Random sense} &
  \multicolumn{1}{l}{Birch} &
  \multicolumn{1}{l}{Jamsic} &
  \multicolumn{1}{l}{BERT KMeans} &
  \multicolumn{1}{l}{Egvi} &
  \multicolumn{1}{l}{BERT KMeans + Egvi} &
  \multicolumn{1}{l}{№ of clusters}\\
\toprule
\rusword{Бог} (God)           & 0.00  & -0.02 & -0.04 & 0.13  & -0.04   &   -0.04   &   0.01 & 3\\
\rusword{Время} (time)        & 0.00  & -0.03 & 0.02  & 0.00  & -0.07   &   -0.01   &   -0.04 & 6\\
\rusword{Год} (year)          & 0.00  & 0.06  & -0.03 & 0.01  & -0.02   &   -0.04   &    0.13  & 3\\
\rusword{Голова} (head)       & 0.00  & -0.03 & 0.51  & -0.02 & 0.20    &   0.61    &   -0.02    & 4 \\
\rusword{Город} (city)        & 0.00  & -0.01 & -0.03 & 0.01  & -0.01   &   1.00    &    0.00   & 2 \\
\rusword{Государство} (state) & -0.06 & 0.02  & -0.04 & -0.06 &  -0.05  &   -0.04   &   -0.05       & 3\\
\rusword{Дело} (business)     & 0.00  & 0.03  & -0.01 & 0.00  & 0.01    &   0.02    &   0.08    & 11 \\
\rusword{День} (day)          & 0.00  & 0.11  & -0.05 & 0.08  & -0.02   &   -0.05   &   0.00    & 5\\
\rusword{Друг} (friend)       & 0.00  & -0.02 & 0.25  & -0.04 & -0.09   &   0.12    &   -0.01    & 3 \\
\rusword{Жена} (wife)         & 0.00  & -0.03 & 0.00  & -0.03 & -0.04   &   0.00    &    0.0    & 2 \\
\rusword{Женщина} (woman)     & 1.00  & 0.00  & 0.00  & 0.00  & 0.00    &   1.00    &    1.0   & 1 \\
\rusword{Жизнь} (life)        & 0.00  & 0.00  & 0.33  & -0.01 & -0.13   &   -0.04   &   0.04    & 4 \\
\rusword{Лицо} (face)         & 0.00  & -0.02 & 0.05  & 0.50  & 0.39    &   0.00    &   0.61    & 3 \\
\rusword{Место} (place)       & 0.00  & 0.05  & -0.10 & 0.09  & 0.01    &   -0.04   &   -0.02   & 4\\
\rusword{Мир} (world)         & 0.00  & -0.01 & 0.13  & 0.14  & 0.04    &   0.20    &   0.00    & 5 \\
\rusword{Ночь} (night)        & 1.00  & 0.00  & 0.00  & 0.00  & 0.00    &   0.00    &   0.00    & 1 \\
\rusword{Работа} (work)       & 0.00  & -0.01 & -0.04 & -0.04 & -0.04   &   0.15    &   0.05    & 5 \\
\rusword{Результат} (result)  & 0.00  & 0.01  & -0.07 & 0.04  & 0.01    &   0.63    &  -0.03     & 2 \\
\rusword{Рука} (hand)         & 0.00  & 0.00  & -0.08 & 0.06  & 0.38    &   0.16    &   0.05    & 3 \\
\rusword{Сила} (power)        & 0.00  & -0.01 & -0.02 & -0.02 & 0.22    &   -0.02   &   0.04   & 6\\
\rusword{Слово} (word)        & 0.00  & -0.03 & 0.01  & 0.01  & 0.19    &   0.00    &   0.00    & 3 \\
\rusword{Сторона} (side)      & 0.00  & -0.01 & -0.04 & 0.21  & 0.30    &   0.41    &   0.20    & 5 \\
\rusword{Тысяча} (thousand)   & 0.00  & 0.00  & -0.08 & -0.01 & -0.05   &   -0.04   &   -0.01    & 3 \\
\rusword{Человек} (human)     & 0.00  & 0.00  & -0.04 & -0.06 & -0.00   &   0.00    &   0.00    & 3 \\
\bottomrule
\end{tabular}
\label{detailed_wsi}
}
\caption{Detailed performance of WSI methods.}
\end{table*}

\end{document}